# Predictive Modeling of ICU Healthcare-Associated Infections from Imbalanced Data. Using Ensembles and a Clustering-Based Undersampling Approach

**Fernando Sánchez-Hernández [1], Juan Carlos Ballesteros-Herráez [2], Mohamed S. Kraiem [3], Mercedes Sánchez-Barba [4] and María N. Moreno-García [3,\***

1. Faculty of Nursing and Physiotherapy, University of Salamanca, Salamanca, 37007, Spain; fsh@usal.es
2. Intensive Care Unit, University Hospital of Salamanca, Salamanca, 37007, Spain; jcballes1969@gmail.com
3. Department of Computing and Automation, University of Salamanca, Salamanca, 37008, Spain; ing_kriem@yahoo.com
4. Department of Statistics, University of Salamanca, Salamanca, 37007, Spain; mersanbar@usal.es
* Correspondence: mmg@usal.es

**Abstract:** Early detection of patients vulnerable to infections acquired in the hospital environment is a challenge in current health systems given the impact that such infections have on patient mortality and healthcare costs. This work is focused on both the identification of risk factors and the prediction of healthcare-associated infections in intensive-care units by means of machine-learning methods. The aim is to support decision making addressed at reducing the incidence rate of infections. In this field, it is necessary to deal with the problem of building reliable classifiers from imbalanced datasets. We propose a clustering-based undersampling strategy to be used in combination with ensemble classifiers. A comparative study with data from 4616 patients was conducted in order to validate our proposal. We applied several single and ensemble classifiers both to the original dataset and to data preprocessed by means of different resampling methods. The results were analyzed by means of classic and recent metrics specifically designed for imbalanced data classification. They revealed that the proposal is more efficient in comparison with other approaches.

## 1. Introduction

Healthcare-associated infections (HAI) are one of the major problems of health systems in many countries due to their direct impact on morbidity, mortality, length of hospital stays, and costs [1]. According to a CDC (Centers for Disease Control and Prevention) report, the estimated overall annual direct medical cost of HAI in U.S. hospitals ranges between $28.4 and $45 billion, and the deaths they cause amount to more than 98,000. It is also estimated that 20% of infections are preventable and the financial benefits of prevention in U.S. hospitals range from $5.7 to $31.5 billion. In addition, another study has found that the benefits of mortality risk reductions are at least 5 times greater than the benefits of only reducing direct medical costs that emerge in hospitals [2]. Detection and surveillance systems are crucial for the timely implementation of appropriate preventive measures.

A significant percentage of HAIs occur in intensive-care units (ICUs), where patients are usually more susceptible to acquiring infections, which results in higher mortality rates or longer stays [3].

Most HAIs diagnosed in ICUs are device-associated infections, which have a great impact on patient progress. They are caused by invasive devices that alter natural defense barriers and favor the transmission of pathogens, which often have high rates of antimicrobial resistance and are part of the ICU flora.

The incidence rate of infections in the ICU has been reduced by means of the implementation of surveillance and prevention measures. In this context, effective decision making implies managing a significant quantity and variety of information, which is usually time-consuming for physicians. Machine-learning techniques can provide support for data processing, not only to automate and improve the efficiency of the decisions made, but also to find valuable patterns in the data that cannot be discovered with alternative procedures. Nowadays, there is a lack of automatic systems for surveillance and diagnosis in this field, and most of them either make use of basic machine-learning methods or require human intervention to introduce domain knowledge [4]. In order to fill this gap, we propose an approach for obtaining reliable HAI predictive models from imbalanced datasets, which allow clinicians to automatically detect the patients most susceptible to infections.

In this work, machine-learning techniques are used to detect the most important HAI risk factors and to identify patients who are more susceptible to infections, taking into account their characteristics, treatment, invasive devices used and other information concerning their stay in the ICU. The study draws upon data from 4616 patients, gathered in the ICU of the University Hospital of Salamanca (Spain) over seven years. This dataset presents an acute imbalance regarding patient classification, since only 311 of the patients analyzed, which represent 6.7% of the total, contracted infections. The application of classification algorithms to imbalanced datasets, such as the one involved in this study, has serious weaknesses, since the achievement of good global accuracy does not equate to precision for the minority class.

Classification from imbalanced datasets represents an important obstacle in supervised learning. It occurs when there is a big difference between the number of instances of each class under study. In these situations, the precision for the minority class is usually significantly lower than the precision for the majority class; therefore, predictive models are not valid even when they present an acceptable accuracy. As such, the classifier can achieve a high percentage of correctly classified instances, but the percentage of instances belonging to the minority class that are correctly classified can be very low. Additionally, the minority class is usually the most interesting one in terms of the application domain; therefore, misclassification of its instances is the type of error with the greatest negative impact on decision-making.

There are several methods to deal with this problem. They can be organized in the following categories [5]:

- Data resampling: oversampling or undersampling procedures to modify the training set by creating or eliminating instances in order to obtain a balanced distribution of the instances belonging to each class.
- Algorithmic modification: this involves the modification of the learning algorithms to make them more suitable for processing imbalanced data.
- Cost-sensitive learning: this takes into account the misclassification costs, so that the different types of errors are treated in a different way.

Sampling strategies are the most widely used. Oversampling the examples of the minority class and under-sampling those belonging to the majority class are two common preprocessing methods applied to deal with imbalanced datasets, but they have some well-known weaknesses. Many works in the literature propose different ways to implement these methods, and there are also some comparative studies about the performance of different resampling strategies [6], as well as about the evaluation of oversampling versus undersampling. Despite the fact that some of the studies yield contradictory results [7], most authors agree on the shortcomings associated with these approaches. Removing potentially valuable data is the main drawback of undersampling the majority class, while oversampling the minority class can cause both overfitting problems and an increase in the computational cost of inducing the models [8]. On the other hand, classification models, as decision trees induced from the oversampled datasets, are usually very large and complex [9].

The other approaches that deal with the problem of imbalanced data, algorithmic modification and cost-sensitive learning, are less often employed owing to certain difficulties in their application. Adapting every algorithm to imbalanced data demands a great effort and sometimes provides poorer results than resampling techniques. In addition, cost-sensitive learning usually requires domain experts to provide values to fill in the cost matrix containing the penalties for the different types of misclassification [5], which is a difficult endeavor.

Regarding classification algorithms, some works have proved that the ensemble approaches usually present the best behavior when working with imbalanced data. These algorithms are often used to induce more reliable classifiers, using either unsampled or resampled datasets as input [10–12]. This subject will be discussed in more detail in subsequent sections.

The aim of the present study is to propose a way of building reliable predictive models from imbalanced datasets, minimizing the negative effects of sampling strategies. The validation is performed through an empirical study in which, besides classical metrics, other novel metrics designed for imbalanced data classification are also applied. The results prove that a suitable combination of ensemble classifiers and controlled undersampling yields better results than other recognized methods.

The rest of the paper is organized as follows: Some important work in the literature about the topic under study is described in Section 2. Section 3 is devoted to introducing background information about the validation and classification approaches, as well as to presenting the proposed method. Specifically, the Section 3.1 presents the validation metrics used in imbalanced data contexts. The Section 3.2 includes the basis of the classification algorithms used in this work, with special emphasis on ensemble classifiers. The presentation and the rationale of the proposed approach are given in Section 3.3. The experimental study and its results are reported in Section 4, and the discussion of these results is to be found in Section 5. Finally, the conclusions are discussed in the last section.

## 2. Related Work

The classification from imbalanced datasets has been the target of intensive research for many years. As mentioned in the introduction, there are different ways of dealing with this drawback, although the most extended involve data preprocessing by means of resampling strategies.

The synthetic minority oversampling technique (SMOTE) is a widely used oversampling method for creating artificial instances of the minority class by introducing new computed examples along the line segments joining the k minority class nearest neighbors. Several extensions of SMOTE have been developed with the aim of improving the precision of the minority class under different circumstances. Some of them involve the creation of instances performed by SMOTE to specific parts of the input space within the limits of the positive class [13,14], while others use noise filters after the application of SMOTE. Sáez et al. [15] proposed the SMOTE-IPF resampling method, which extends SMOTE with an ensemble-based noise filter called the iterative-partitioning filter (IPF); the said filter can overcome the problems produced by noisy and borderline examples in imbalanced datasets.

Recently a neural network-based oversampling approach has been used in some works in the literature to generate minority class instances. In these works, unlike SMOTE-based methods that do not take into account the data distribution, generative adversarial networks (GAN) algorithms are used for learning class-dependent data distribution and produce the generative model [16]. These techniques are mainly applied in the image processing area, where several GAN variants have been implemented in order to generate synthetic images [17]. Although the results are promising, the main drawback of this approach in comparison with classic oversampling strategies stems from its higher complexity and computational time, since a time-consuming training process is required to induce the models. Moreover, GAN methods are not free from the overfitting problem.

Another popular sampling approach is random under-sampling (RUS). This technique balances the examples of the dataset by randomly removing some of them from the majority class. Its main drawback is the elimination of data that could be important for inducing the classification model [9].

Condensed nearest neighbor (CNN) [18] is a data-reduction strategy that was not initially designed for dealing with imbalanced classification but for improving the efficiency of the nearest neighbor classification algorithm. CNN identifies prototype points from the original dataset to be used to classify new instances. It uses an iterative procedure, which starts with a subset containing a prototype point randomly selected. Then, the remaining points in the dataset are classified by the NN rule using the points in the prototype subset. Those that are classified incorrectly are added to the subsets. A similar proposal put forward by Tomek [19] adds a pre-processing step involving the removal of noise and borderline examples in order to obtain further data reduction without increasing the classification error rate. Edited nearest neighbor (ENN) [20] is a similar strategy that includes a preliminary pass to remove points of the majority class. These are the examples whose neighbors belong mostly to the other class.

NearMiss [21] is a family of methods where the majority class is undersampled by means of a procedure meant to remove examples based on their distance to other examples of the same class. NearMiss-1 retains the majority class samples which are close to some minority class samples, since their average distances to the nearest minority class samples are the smallest. NearMiss-2 selects the majority class samples for which the average distances to the kfarthest minority class samples are the smallest. For each minority sample, NearMiss3 selects the k closest majority class samples.

One-sided selection (OSS) is also a well-known undersampling technique [22] that uses k-NN to classify the instances of a subset of the training set initially containing all the examples of the minority class and randomly selected examples of the majority class. The majority class instances that are correctly classified are discarded because they are considered redundant. Noise and borderline instances are also removed. Due to its questionable effectiveness, especially when the imbalance rate is high, OSS is not usually used alone, but in combination with oversampling strategies [6]. There are some proposals in the literature that are mainly focused on minimizing the information loss caused by under-sampling. In [23], majority-class examples are separated into clusters that are later used to build different training subsets. Each subset is formed for all the minority class instances and as many majority class samples as clusters. These samples are proportional to the size of the clusters they belong to. The final output is given by the aggregation of the predicted results of individual classifiers.

Since SVM (support vector machines) are one of the most suitable algorithms for binary classification, especially for high dimensionality problems, they have been extensively studied in imbalanced data contexts where they do not always have the desired behavior. However, an SVM can be adapted to generate an asymmetric margin, wider at the side of the minority class, in order to increase its performance [24,25]. In [26], an evaluation of different well-known strategies for imbalanced text classification using SVM was carried out. The results showed that resampling and weighting strategies were not effective in that application domain. Soft-margin polynomial SVM combined with resampling methods have shown good performance in a study for classifying medical documents in which Unified Medical Language Systems (UMLS) were used for medical terms extraction [27].

Ensemble methods have also been applied to learn from imbalanced datasets, both alone and combined with other kinds of data sampling [5,10,11]. In [28], the imbalance problem is addressed by means of techniques originally created to increase the ensemble diversity. Some of the methods for the promotion of diversity expand the feature space with new attributes, but most of them are also based on sampling strategies. The study shows some improvement when using diversity techniques, but the results can be generalized neither to all problems considered nor for all quality measures. Diversity measures are used to prune base classifiers in the ensemble models in order to avoid overfitting, but the effects of diversity on accuracy are unclear. In addition, the pruning process requires the time-consuming task of training and evaluating many classifiers, which is a noteworthy drawback of these approaches [29]. One concern when using ensembles is the interpretability of the output, which becomes difficult due to the complexity of the models generated. In order to make them more understandable, a method is proposed in [30] that provides a single set of production rules by combining and simplifying the output of an ensemble of binary decision trees, without serious effects on performance.

The problem of imbalanced data is very common in the medical field and has been addressed in some works, such as studies of mortality [12,31,32], treatment outcomes [33], drug toxicity assessment [34] and medical diagnosis [35–37]. Preliminary studies about the behavior of ensemble classifiers, as opposed to single classifiers in imbalanced data contexts, are conducted to predict the mortality of polytraumatized patients [12] and the success of non-invasive mechanical ventilation [33]. These works proved the good behavior of ensembles, even when applied to high-dimensionality datasets preprocessed with feature selection methods.

In [38], several machine-learning algorithms are applied to predict the outcome of the implantation of individual embryos in in-vitro fertilization. The implantation cases in the dataset were far fewer than the no-implantation ones; thus, the authors address the great imbalance of data simply by building the training and the test sets with the same proportion of cases of the minority and the majority class as in the original dataset. However, no further treatment is performed. The impact of SMOTE on the performance of three well-known classifiers (probabilistic neural network, naïve Bayes, and decision tree) is analyzed in a study concerning diabetes prediction [39]. For the three algorithms, sensitivity increased when the dataset was widely oversampled, but as expected, accuracy and specificity decreased in all cases.

Work specifically aimed at the study of healthcare-associated infections is scarcer. There are works in the literature in which different machine learning techniques are applied to predict infections but in most of them the problem of imbalanced data is not addressed. One of these works is presented in [4], where a case-based reasoning (CBR) model is used to make automatic diagnostics of HAIs. The system, which includes expert-defined and automatically generated rules, achieve an accuracy degree similar to that of the experts. Therefore, it is required domain knowledge to obtain some of the rules, which is a major drawback

Resampling strategies for generating synthetic instances were applied in a prevalence study of nosocomial infections [40]. It involves another strategy based on asymmetrical soft margin for SVM. The purpose of the study is the automatic identification of patients with a high risk of acquiring HAIs, facilitating in this way the time-consuming task of infection surveillance and control. The results prove that both approaches improve sensitivity values, but receiver operator characteristic (ROC) analysis is not performed. A more recent paper [41] analyses the incidence and risk factors of healthcare-associated ventriculitis and meningitis in a neuro-ICU by means of tree-based machine learning algorithms. The problem of imbalanced data is addressed by preprocessing data with the SMOTE resampling technique. The work is mainly focused on analyzing risk factors and feature selection. The results showed a better performance of tree-based algorithms than regression models. Another additional advantage pointed out by the authors is the fact that the first methods allow the identification of interaction between factors, and that information can be used to take preventive measures. Risk factors selection was improved by combining the results from relative risk analysis, regression and machine learning methods.

After analyzing different approaches to address the problem of classification from imbalanced data, we can conclude that there is no fully satisfactory and widely applicable solution. Each strategy has its advantages and disadvantages, and their respective effectiveness depends on multiple factors. Oversampling methods are characterized by the problem of overfitting, which reduces the applicability of the classification model. Some proposals have been made to improve SMOTE, one of the most popular oversampling strategies, through the treatment of noise and borderline examples, but the overfitting drawback persists. Generative adversarial networks have emerged as new and promising oversampling techniques, although their application is practically limited to the field of image processing. Their high computational cost is one of their main drawbacks. Undersampling techniques do not are affected by overfitting problems, but rather from a loss of information due to the elimination of instances. The improvements with respect to the basic method, RUS, also focus on the treatment of borderline examples and noise, as well as on the selection of the most representative points of the majority class. These strategies generally perform worse than oversampling; therefore, they are often applied in combination. Finally, ensemble classifiers are an alternative to sampling strategies since they consist of classifiers induced from different hypothesis spaces. Although most of them also use resampling, the overfitting problem is not as pronounced as in oversampling methods due to the ensemble diversity and the fact that resampling does not focus on instances of a particular class. Their performance is usually good, although variable relative to the application domain. Their higher complexity and longer model induction time are their main shortcomings against single classifiers.

## 3. Materials and Methods

In this work, several supervised learning algorithms, specifically classification algorithms, have been applied to evaluate our proposal. Its validation has been carried out using traditional metrics such as accuracy or precision, as well as classic and new metrics designed specifically to evaluate classification models induced from imbalanced data. These approaches, along with the proposed method, are described in the following sections.

*3.1. Validation of Classifiers in Imbalanced Data Contexts*

Cross-validation is the method most employed for validating classifiers. It is an effective procedure for approximating the errors that might occur when a classifier is used to unlabeled data. By means of k-fold cross-validation, the available data are divided into k disjoint subsets of the same size. The k trainings are performed by taking in each of them a different subset as test set and building the model with the remainder of the sub-sets. The error rate is the average of the errors obtained after testing the different classifiers induced from the k training sets.

In many research works, the validation of classifiers is carried out only by means of examining their accuracy, that is, the percentage of correctly classified instances. However, this measure is not appropriate in imbalanced data contexts because in these scenarios, machine-learning algorithms can achieve an acceptable global accuracy, but the precision for the minority class can be very low. Therefore, accuracy can be complemented with other metrics that provide additional error perspectives, especially when evaluating binary decision problems. In these cases, the examples are classified as either positive or negative, and the output of the classifier can belong to one of the following four categories: true positives (TP), positive instances correctly classified; false positives (FP), negative instances classified as positive; true negatives (TN), negative instances correctly classified; and false negatives (FN), positive instances classified as negative. Given this information, it is possible to define certain validation metrics such as precision, recall, F-measure or area under the ROC curve.

Precision is the probability of an example being positive if the classifier classifies it as positive:

$$Precision = TP / (TP + FP) \tag{1}$$

Recall or sensitivity refers to the probability of a positive example being classified as positive.

$$Recall = TP / (TP + FN) \tag{2}$$

A good classifier will provide high recall and precision values; however, when one of them increases, the other one often decreases. For this reason, it can be very difficult to achieve high values for both parameters simultaneously. When working with imbalanced datasets, the objective is to improve the recall without worsening the precision [7]. A metric that combines precision and recall is the F-measure.

$$F - measure = \frac{(1 + \beta 2) * Precision * Recall}{\beta 2 \, Precision + Recall} \tag{3}$$

where $\beta$ represents the relative importance of precision and recall, and its value is usually set to 1.

The ROC curve is a well-known approach for evaluating the classifier performance for different values of TPR (true positive rate) and FPR (false positive rate). The ROC curve is the representation of TPR against FPR.

$$TPR = TP / (TP + FN) \tag{4}$$

$$FPR = FP / (FP + TN) \tag{5}$$

Point (0,0) of the ROC graph corresponds to a classifier that classifies all examples as negative, and point (1,1) to a classifier that classifies all examples as positive. The area under the ROC curve (AUC) is a robust method to identify optimal classifiers. The best learning system will be the one that provides a set of classifiers with a greater area under the ROC curve.

The above metrics provide more insight than the single accuracy about the error of classifiers concerning the instances of each class, but they have not been specifically defined for imbalanced data classification. However, G-mean, the geometric mean of TPR and TNR (true negative rate), has been shown to be indicative of this type of problems. This metric considers the correct classification of instances for both positive and negative classes.

Some authors argue that AUC and G-mean share the drawback of not differentiating the contribution of each class to the overall accuracy. In [42], this problem is addressed by proposing a new metric called optimized precision (*OP*), which is defined as follows:

$$OP = \frac{Accuracy - |TNR - TPR|}{TNR + TPR} \tag{6}$$

*OP* takes the optimal value for a given overall accuracy when the true negative rate and true positive rate are very close to each other.

### 3.2. Classification Algorithms

In this study, several classification algorithms were applied. They were both simple and ensemble classifiers. As simple classifiers, we tested decision trees, Bayesian networks and SVM, although the results provided for the two last were remarkably poor and have not been reported. The ensembles applied were Random Tree, Random Forest, Bagging, AdaBoost and Random Committee. In addition, some neural network algorithms were tested, but they yielded worse results than simple classifiers, with the added drawback of the much longer computational time.

Ensembles belong to the category of multiclassifiers, which combine several individual classifiers induced with different basic methods or obtained from different training datasets, with the aim of improving the accuracy of the predictions. They can be divided into two groups. The first, also named ensemble classifiers, such as bagging [43], boosting [44] and random forest [45], induce models that merge classifiers with the same learning algorithm, but introduce modifications in the training data set. The second type of methods, named hybrids, such as stacking [46] and cascading [47], create new hybrid learning techniques from different base-learning algorithms.

Bagging is the acronym for Bootstrap AGGregatING. The method induces a multiclassifier that consists of an ensemble of single classifiers built on bootstrap replicates of the training set. Each classifier in the ensemble is trained with a set of examples generated randomly with replacement from the original training set, so that some examples may be repeated. Bagging uses a majority vote to combine the outputs of the classifiers in the ensemble. This procedure is an abstract level method, in which no information about the probability or correctness of the predicted labels is known. By contrast, other approaches, such as rank level or measurement level methods, provide rank and confidence information, respectively [48].

Boosting is a multiclassifier of the same kind as Bagging; however, this method assigns weights to the outputs of the induced single classifiers from different training sets (strategies). The weight of a strategy represents the probability of it being the most accurate of all. In an iterative process, the weights are updated by increasing the weight of strategies with the correct prediction and reducing the weight of strategies with incorrect predictions. In this way, the multiclassifier is developed incrementally, adding one classifier at a time. The classifier that joins the ensemble at step k is trained on a data set selectively sampled from the training data set Z. The sampling distribution starts from uniform and progresses in each k step towards increasing the likelihood of worst classified data points at step k − 1. This algorithm is called AdaBoost and comes from ADAptive BOOSTing. This algorithm presents the advantage of driving the ensemble training error to zero in very few iterations [48].

Random forest [45] is an algorithm widely used in medical fields [49] that induces many decision trees, called random trees, each of which produces its own output for a given unclassified example. The most popular class obtained by simple vote is chosen as the final outcome. It is a variant of bagging where the induction of each tree is produced from a bootstrap sample obtained from the original data set, independently taking n examples with replacement, but with the same distribution for all trees in the forest. About one-third of the examples not used to build a tree, the out-of-bag (OOB) sample, are used as test set. Moreover, the (random) trees are built from a randomly selected subset S of M features from the original dataset of N features. The selection is carried out at each node of the tree, then the best feature in S to split the node is searched. The value of M suggested by Breiman is $[\log_2 N + 1]$. In this way, CART (Classification and Regression Trees) is fully grown without pruning.

The random committee algorithm is used to induce an ensemble where each base classifier is built using a different random number of seeds. The final prediction is generated by averaging probability estimations over the individual base classifiers.

### 3.3. Addressing Imbalanced Data Classification

The aim of this study is to induce predictive models that can be used to identify patients with high risk of HAIs in the ICU. As mentioned before, the main problem to be addressed in order to obtain reliable classification models is the treatment of imbalanced data. If used individually on imbalanced datasets, classifiers sometimes fail, yielding a very low precision in the classification of minority class examples. This work provides a proposal to deal with this drawback while avoiding some of the disadvantages of the usual approaches. Specifically, the aim is to give reliable alternatives to the simple use of sampling strategies that focus on a particular class, either undersampling the majority class or oversampling the minority one.

The proposal is based on identifying regions in the space of characteristics with different imbalance degrees in order to reduce the area of resampling and its negative effects. Since oversampling strategies cause overfitting, which is one of the most adverse consequences of resampling, our procedure involves undersampling examples of the majority class. However, only regions exceeding a threshold of imbalance ratio are undersampled in order to minimize the loss of information. This approach is combined with the use of ensemble classifiers, which build several hypotheses from different datasets following a resampling strategy, too. Nevertheless, this strategy differs from that used in the classical treatment of imbalanced problems because it is not focused on one particular class; as such, the outcome is not biased in favor of a specific class of instances. In addition, ensemble methods have the potential capacity to minimize overfitting problems. This fact is essential when dealing with imbalanced data, since, as stated before, this is one of the main weaknesses of oversampling.

#### 3.3.1. Rationale for the Choice of Ensemble Classifiers

According to Kuncheva [48], the key to the good behavior of classifier ensembles is the diversity provided by different training sets. The ideal classifier should be induced from a training set randomly generated from the distribution of the entire space of hypothesis. However, only one training set $Z = \{z_1 \dots z_N\}$ is usually available and does not have an appropriate size to be split into mutually exclusive subsets with a considerable number of instances. In these cases, the bootstrap sampling can be used to generate $L$ training sets. When training sets are generated from Bootstrap sampling, significant improvements are achieved mainly if the base classifier is unstable, that is, small changes in the training set should lead to large changes in the classifier output. Although instability depends on the learning algorithm, it is also influenced by the dataset characteristics, such as data imbalance.

On the other hand, majority vote properties usually ensure the improvement of the single classifier results if the outputs are independent and the individual accuracies of the base classifiers have a normal distribution. This fact was evidenced in [48] by means of the rationale described below.

Given a set of labels $\Omega = \{\omega_1, \ldots, \omega_c\}$, a set of classifiers $D = \{D_1, \ldots, D_L\}$ and objects $x = [x_1, \ldots, x_n]^T \in \mathcal{R}^n$ to be classified, each classifier $D_i$ in the ensemble provides a label $s_i \in \Omega, i = 1, \ldots, L$. Thus, for any object $x \in \mathcal{R}^n$ to be classified, the $L$ classifier outputs define a vector $s = [s_1, \ldots, s_L]^T \in \Omega^L$. The labels given by the classifiers can be represented as binary vectors $[d_{i,1}, \ldots, d_{i,c}]^T \in \{0,1\}^c, i = 1, \ldots, L$, where $d_{i,j} = 1$ if the classifier $D_i$ assigns to **x** the label $\omega_j$, and 0 otherwise. The final choice of the class is obtained by majority vote. This means that the following rule must be complied with:

$$\sum_{i=1}^{L} d_{i,k} = max_{j=1}^{c} \sum_{i=1}^{n} d_{i,j} \tag{7}$$

We must take into account that the number of classifiers, *L*, must be odd for binary classifications. Let's consider that the probability of predicting the right class for any $x \in \mathcal{R}^n$ for each classifier is *p*. If the classifier outputs are independent, then the joint probability for any subset of classifiers $A \subseteq D, A = \{D_{i1}, \ldots, D_{ik}\}$ can be decomposed as $P(D_{i1} = s_{i1} \ldots, D_{ik} = s_{ik}) = P(D_{i1} = s_{i1}) \times \ldots \times P(D_{ik} = s_{ik})$, where $s_{ij}$ is de label given by the classifier $D_{ij}$.

When using majority vote, the ensemble output will be correct if at least [L/2] + 1 classifiers give the correct label. When the individual probabilities *p* are the same, the accuracy of the ensemble is:

$$P = \sum_{m=\left[\frac{L}{2}\right]+1}^{L} \binom{L}{m} p^m (1-p)^{L-m} \tag{8}$$

For unequal probabilities, when the distribution of the individual probabilities $p_i$ is normal, the value of *p* can be substituted for the mean value of $p_i$ in the equation.

According to the Condorcet theorem, if the individual accuracies $p > 0.5$ then $P \to 1$ as $L \to \infty$. Therefore, in many problems, the use of majority vote-based ensembles, such as bagging and random forest, is expected to lead to an improvement in individual accuracy even if the base classifiers are not completely independent.

In addition to accuracy, there are other aspects that need to be studied when using ensemble classifiers, one of which is overfitting. This problem, associated with data oversampling, causes loss of generalization capacity in classifiers when they are very adapted to the training set and show poor predictive performance for classifying other examples. This shortcoming can originate from the complexity of the classifiers and may be analyzed by means of the bias-variance decomposition of the classification error [50,51]. Although there are different definitions and measures of the bias and variance concepts, bias can be generally defined as a measure of the difference between the true and predicted values, while variance is the variability of the predictions of the classifier.

Following the previous notation, if *D* is a classifier randomly chosen, *X* is a point in the feature space and $\omega \in \Omega$ is the class label of *X*, the posterior probabilities for the *c* classes given *X* are $P(\omega_i|X)$ and the probabilities for any possible classifier *D* are $P_D(\omega_i|X), i = 1, \ldots, c$. Taking into account both probabilities, bias measures are based on comparisons between true probability values $P(\omega_i|X)$ and predicted values $P_D(\omega_i|X)$. Measures of variance are only focused on the probability distribution of the predictions $P_D(\omega_i|X)$.

It is necessary to minimize bias and variance in order to obtain a good performance; however, it is difficult to jointly minimize both values because low bias is usually related to high variance and vice versa. Large bias is characteristic of simple and inflexible models, whereas high variance is typical of very flexible models. High bias is associated with underfitting, which occurs when the classifier is not optimal to classify the examples; however, high variance can cause overfitting. Some causes of the variance are noise in the data, train sample and randomization. In imbalanced datasets, the noise in examples of the minority class has a great impact on the induced models and can be the cause of overfitting. As such, in these cases it is important to resort to methods that avoid this problem.

Ensemble classifiers can also suffer overfitting because they make use of sampling. Several studies about the behavior of ensembles regarding bias and variance have been carried out, and most of them agree on the fact that Bagging reduces variance without increasing bias, which is usually low, too [48,50–53]. The same studies have found that boosting behaves differently over the course of its iterations. Bias is reduced during the first iterations, while the variance is mainly reduced during the last steps. AdaBoost is good at avoiding overfitting problems and reducing errors, even though the complexity of the induced classifiers gradually increases. Boosting and AdaBoost might lose performance on noisy data, especially for small datasets; however, bagging does not have this problem. Random forest can benefit from the advantages of bagging due to their common characteristics. On the other hand, it is admitted that random forest is relatively robust to outliers and noise. This is in part due to the way it generates the samples from both the feature set and the dataset.

All these reasons have guided the proposal presented in this work, which combines the use of ensembles and dataset sampling.

3.3.2. Clustering-Based Random Undersampling

The results of ensemble classifiers applied to imbalanced data can be improved by combining them with resampling strategies. Oversampling usually leads to a better classification performance of the minority class examples if the dataset size is not too large. However, the replication of instances causes overfitting; therefore, classifiers induced from resampled datasets become too adapted to them and less extensible to other data. Undersampling may be a good alternative to avoid this drawback, especially if the dataset is large, but the loss of information associated with the elimination of majority class instances could negatively affect the classification results. The degree of the impact depends to a great extent on the application domain and the dataset characteristics, such as imbalance ratio, number of instances, attributes, etc.

In order to avoid overfitting problems and minimize the adverse effects of undersampling, our approach (clustering-RUS) involves undersampling some regions in the entire space of characteristics. To create those regions, a clustering technique is applied in the $n$ dimensional space, where $n$ is the number of characteristics excluding the class attribute. This ensures that regions are independent of the class and that there are not many more instances belonging to one class than to the other. Since the clusters have a different imbalance ratio, only those exceeding a specified threshold will be undersampled. In this way, the loss of information is minimized because it is only applied to a subset of instances instead of an entire set. The threshold value must represent a significative reduction with respect to the imbalance ratio of the initial dataset. We suggest that the reduction be in an interval between 25% and 30%. To select the number of clusters, it must be checked that at least one of them has an imbalance ratio significantly lower than the established threshold value, starting with two clusters and increasing the number until the condition is fulfilled. Once the clusters have been created and those that need it have been sampled, a classifier per each group of clusters is induced. That is, one for clusters with an imbalance ratio below of the established threshold and another for the remainder clusters. When new instances need to be classified, it is necessary to check what cluster they belong to by computing its distance to the centroids of the clusters and then they are classified by the classifier corresponding to that cluster.

Figure 1 shows all the steps necessary for implementing and validating this approach. The overlapped boxes drawn with dashed lines indicate that the processes enclosed in them are performed once for each fold generated in the cross-validation procedure (.explained in Section 3.1). In the first step of the proposed strategy, a clustering algorithm is applied to the dataset to split it into two subsets with low and high imbalance, respectively. Then, only the training set of high imbalanced clusters is undersampled (lower branch in the figure). The next steps are the application of the classification algorithm and the evaluation of the model performance, which is carried out separately for the two dataset partitions. The final output is the weighted average performance.

In the empirical study carried out in the context of this work, the clusters were created by using the k-means algorithm with the normalized Euclidean distance measure.

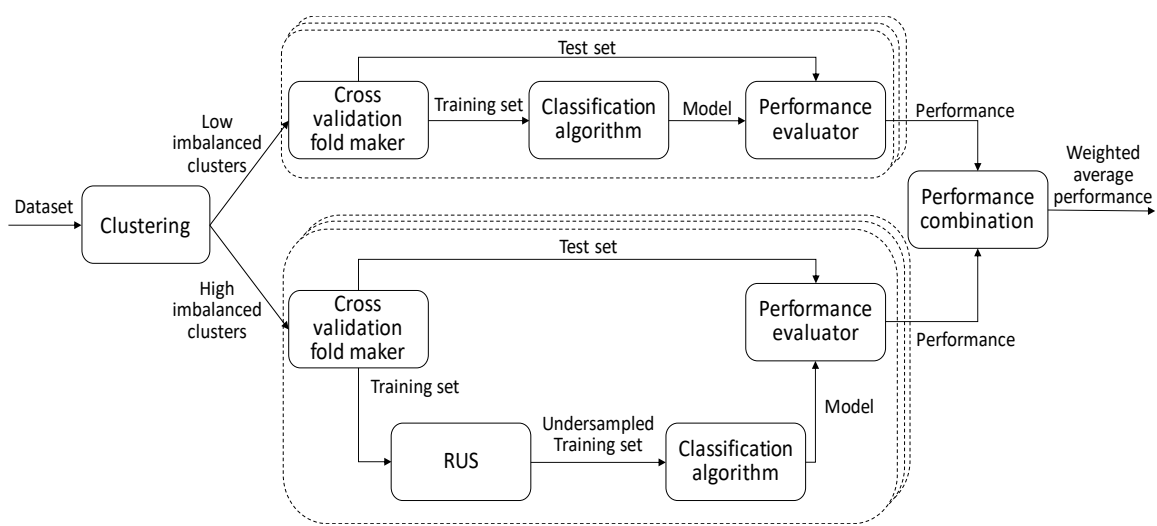

**Figure 1.** Process of inducing and validating classification models when using the clustering based random undersampling approach.

## 4. Results

The clustering-based random undersampling method, proposed in this work to address the problem of imbalanced data classification, was validated through an experimental study conducted in the application domain under research. Thus, data from ICU patients were used to induce the classifiers in charge of predicting healthcare-associated infections. The following Sections focus on providing details of the study and its results.

*4.1. Description of the Empirical Study*

Based on the assumptions stated in Section 3.3.1, several ensemble classifiers were applied to the available dataset, both before and after its processing, using different sampling strategies, including our proposal. We compared the results of the classifiers obtained from preprocessed data using our sampling approach in comparison with those obtained from the original dataset, as well as from the datasets resulting from the application of SMOTE (resampling the minority class 100% and 500%) and RUS sampling methods. RUS was used with a distribution spread of 4 (ratio 4:1 of majority to minority class instances). Since we handle a very imbalanced data set, the RUS ratio should be high enough to improve the classification of the minority class but at the same time as low as possible to lose the minimum information. In our experiments we found that values higher than 4 hardly affect the improvement of the results, so the ratio chosen was 1:4.

The KnowledgeFlow tool of Weka (https://www.cs.waikato.ac.nz) was used to apply the algorithms. All were run with the default parameters.

The dataset used in the study comprises information about 4616 patients hospitalized in the ICU of the University Hospital of Salamanca. We focused on predictive factors of infections; since some attributes such as days of stay or death are not known until the end of the stay, they were discarded. The attributes used in the learning process were the following:

- Gender
- Acute Physiology and Chronic Health Evaluation (APACHE II)
- Emergency surgery
- Immunosuppression
- Neutropenia
- Immunodeficiency
- Mechanical Ventilation

- Central venous catheter (CVC)
- Urinary Catheter
- Parenteral nutrition
- Patient origin
- 48 h of antibiotic treatment
- Previous surgery
- Extrarenal depuration

Year and APACHE are numerical attributes. Origin can take one of four nominal values, while previous surgery can take one of 13 nominal values (no surgery and 12 types of surgery). The rest of the attributes are binary, taking the values YES/NO, except gender, which takes MALE/FEMALE values. The label attribute is infection, which can take the values YES or NO.

Classification algorithms were applied to induce models that make it possible to predict infections in ICU patients. In order to do a comparative study of the results, two kinds of algorithms were used, single classifiers and multiclassifiers, or more specifically, ensemble classifiers. These were tested with several base classifiers. The results of some algorithms providing very poor results (accuracy less than 60%) are not reported. We highlight the fact that, applied in different ways, SVM yielded a minority class precision close to 0%, even though this algorithm has been successfully used for dealing with the problem of imbalanced data in other application domains.

Ten-fold cross-validation was used in the validation of all classifiers. In the experiments carried out with the resampled data using SMOTE, the folders were created before the resampling process in order to ensure that the test set does not contain "synthetic" examples, which could mask the right performance and result in unreal values due to overfitting. The same procedure was used for undersampling.

In the implementation of the clustering-RUS process, two clusters were created, which split the space of characteristics into two regions with imbalance ratios of 6.03 and 17.33. The threshold to perform undersampling was set to 10.0, hence only the cluster with the lower value of imbalance ratio was undersampled. To do that, we also used RUS with a distribution spread of 4.

*4.2. Validation of the Proposal*

In order to validate the proposed method, clustering-based RUS was tested against other resampling strategies by analyzing several quality metrics obtained after applying reliable single and ensemble classifiers. Given the fact that accuracy is not a suitable evaluation measure in imbalanced data contexts, several metrics introduced in Section 3.1 were obtained. These are: accuracy, area under the ROC curve (AUC), weighted average of precision, recall (sensitivity), F-measure, G-mean, and optimized precision (OP). F-measure was computed with the β parameter set to 1.

Table 1 shows the values of these measures obtained by applying the classification algorithms to data that has not been preprocessed with sampling procedures. Tables 2–4 show the same measures, but in this case, they have been obtained by the classifiers induced from data resampled with SMOTE 100%, SMOTE 500% and RUS 4:1, respectively. Finally, Table 5 contains the results obtained when our approach (clustering-RUS) was applied. As previously mentioned, the results of algorithms that yielded very poor results are not included in the tables. Among them are those obtained from some neural networks. We would like to point out that we even tested a deep learning algorithm based on a multi-layer feed-forward artificial neural network with 50 hidden layers. The generated model provided for the original dataset an accuracy of 0.937, a weighted average precision of 0.920 and a weighted average recall of 0.660. We can see in Table 1 that none of the algorithms has such a bad performance, especially when comparing recall since the worst recall value in the table is 0.935.

Table 1. Results of the classifiers without resampling.

| Algorithm | Accuracy | OP | G-Mean | Precision | Recall | F-Measure | AUC |
|---|---|---|---|---|---|---|---|
| J48 | 0.941 | 0.390 | 0.532 | 0.929 | 0.941 | 0.930 | 0.675 |
| Random Forest | 0.949 | 0.563 | 0.656 | 0.942 | 0.949 | 0.944 | 0.861 |
| Random Tree | 0.925 | 0.602 | 0.684 | 0.928 | 0.925 | 0.926 | 0.726 |
| AdaBoost-J48 | 0.943 | 0.604 | 0.686 | 0.939 | 0.943 | 0.941 | 0.832 |
| AdaBoost-Random Forest | 0.950 | 0.568 | 0.660 | 0.944 | 0.950 | 0.944 | 0.823 |
| AdaBoost-Random Tree | 0.944 | 0.599 | 0.682 | 0.938 | 0.943 | 0.940 | 0.840 |
| Bagging-J48 | 0.940 | 0.389 | 0.531 | 0.928 | 0.940 | 0.929 | 0.812 |
| Bagging-Random Forest | 0.949 | 0.500 | 0.610 | 0.941 | 0.949 | 0.941 | 0.874 |
| Bagging-Random Tree | 0.939 | 0.568 | 0.660 | 0.934 | 0.939 | 0.936 | 0.824 |
| Random Committee-Random Forest | 0.950 | 0.568 | 0.660 | 0.944 | 0.950 | 0.944 | 0.871 |
| Random Committee-Random Tree | 0.935 | 0.597 | 0.681 | 0.933 | 0.935 | 0.934 | 0.820 |

Table 2. Results of the classifiers with synthetic minority oversampling technique (SMOTE) resampling (100%).

| Algorithm | Accuracy | OP | G-Mean | Precision | Recall | F-Measure | AUC |
|---|---|---|---|---|---|---|---|
| J48 | 0.933 | 0.408 | 0.546 | 0.546 | 0.921 | 0.933 | 0.925 |
| Random Forest | 0.950 | 0.583 | 0.671 | 0.671 | 0.943 | 0.950 | 0.945 |
| Random Tree | 0.924 | 0.616 | 0.694 | 0.694 | 0.929 | 0.924 | 0.927 |
| AdaBoost-J48 | 0.942 | 0.600 | 0.683 | 0.683 | 0.937 | 0.942 | 0.939 |
| AdaBoost-Random Forest | 0.947 | 0.566 | 0.659 | 0.659 | 0.940 | 0.947 | 0.942 |
| AdaBoost-Random Tree | 0.941 | 0.588 | 0.674 | 0.674 | 0.936 | 0.941 | 0.938 |
| Bagging-J48 | 0.942 | 0.452 | 0.577 | 0.577 | 0.932 | 0.942 | 0.934 |
| Bagging-Random Forest | 0.947 | 0.539 | 0.639 | 0.639 | 0.939 | 0.946 | 0.941 |
| Bagging-Random Tree | 0.940 | 0.603 | 0.685 | 0.685 | 0.940 | 0.938 | 0.938 |
| Random Committee-Random Forest | 0.949 | 0.579 | 0.668 | 0.668 | 0.942 | 0.949 | 0.944 |
| Random Committee-Random Tree | 0.940 | 0.607 | 0.687 | 0.687 | 0.936 | 0.940 | 0.938 |

Table 3. Results of the classifiers with smote resampling (500%).

| Algorithm | Accuracy | OP | G-Mean | Precision | Recall | F-Measure | AUC |
|---|---|---|---|---|---|---|---|
| J48 | 0.932 | 0.480 | 0.598 | 0.924 | 0.932 | 0.927 | 0.793 |
| Random Forest | 0.943 | 0.573 | 0.663 | 0.937 | 0.943 | 0.939 | 0.865 |
| Random Tree | 0.916 | 0.597 | 0.681 | 0.925 | 0.916 | 0.920 | 0.720 |
| AdaBoost-J48 | 0.936 | 0.604 | 0.686 | 0.934 | 0.936 | 0.935 | 0.830 |
| AdaBoost-Random Forest | 0.946 | 0.588 | 0.674 | 0.940 | 0.946 | 0.942 | 0.814 |
| AdaBoost-Random Tree | 0.939 | 0.584 | 0.671 | 0.934 | 0.936 | 0.936 | 0.828 |
| Bagging-J48 | 0.939 | 0.505 | 0.615 | 0.931 | 0.939 | 0.934 | 0.849 |
| Bagging-Random Forest | 0.942 | 0.540 | 0.640 | 0.935 | 0.942 | 0.937 | 0.873 |
| Bagging-Random Tree | 0.936 | 0.619 | 0.697 | 0.934 | 0.936 | 0.935 | 0.841 |
| Random Committee-Random Forest | 0.946 | 0.575 | 0.664 | 0.939 | 0.946 | 0.941 | 0.873 |
| Random Committee-Random Tree | 0.934 | 0.591 | 0.676 | 0.932 | 0.934 | 0.933 | 0.817 |

Table 4. Results of the classifiers with random under-sampling (RUS) (4:1).

| Algorithm | Accuracy | OP | G-Mean | Precision | Recall | F-Measure | AUC |
|---|---|---|---|---|---|---|---|
| J48 | 0.914 | 0.532 | 0.634 | 0.919 | 0.914 | 0.916 | 0.740 |
| Random Forest | 0.909 | 0.661 | 0.725 | 0.928 | 0.909 | 0.917 | 0.865 |
| Random Tree | 0.868 | 0.700 | 0.747 | 0.924 | 0.868 | 0.890 | 0.760 |
| AdaBoost-J48 | 0.888 | 0.679 | 0.735 | 0.925 | 0.888 | 0.903 | 0.845 |
| AdaBoost-Random Forest | 0.909 | 0.647 | 0.715 | 0.927 | 0.909 | 0.916 | 0.814 |
| AdaBoost-Random Tree | 0.891 | 0.649 | 0.715 | 0.923 | 0.891 | 0.904 | 0.836 |
| Bagging-J48 | 0.907 | 0.597 | 0.680 | 0.922 | 0.907 | 0.914 | 0.841 |
| Bagging-Random Forest | 0.913 | 0.649 | 0.717 | 0.928 | 0.913 | 0.919 | 0.870 |
| Bagging-Random Tree | 0.884 | 0.685 | 0.740 | 0.925 | 0.884 | 0.901 | 0.832 |
| Random Committee-Random Forest | 0.909 | 0.652 | 0.719 | 0.927 | 0.909 | 0.917 | 0.870 |
| Random Committee-Random Tree | 0.883 | 0.666 | 0.726 | 0.923 | 0.883 | 0.899 | 0.843 |

**Table 5.** Results of the classifiers with clustering-RUS (4:1).

| Algorithm | Accuracy | OP | G-Mean | Precision | Recall | F-Measure | AUC |
| --- | --- | --- | --- | --- | --- | --- | --- |
| J48 | 0.908 | 0.502 | 0.614 | 0.915 | 0.908 | 0.911 | 0.704 |
| Random Forest | 0.912 | 0.668 | 0.730 | 0.929 | 0.929 | 0.919 | 0.858 |
| Random Tree | 0.886 | 0.670 | 0.729 | 0.924 | 0.886 | 0.902 | 0.825 |
| AdaBoost-J48 | 0.890 | 0.664 | 0.725 | 0.924 | 0.890 | 0.904 | 0.831 |
| AdaBoost-Random Forest | 0.916 | 0.659 | 0.724 | 0.930 | 0.916 | 0.922 | 0.817 |
| AdaBoost-Random Tree | 0.886 | 0.670 | 0.729 | 0.924 | 0.886 | 0.902 | 0.825 |
| Bagging-J48 | 0.914 | 0.549 | 0.646 | 0.920 | 0.914 | 0.917 | 0.844 |
| Bagging-Random Forest | 0.920 | 0.649 | 0.717 | 0.930 | 0.920 | 0.924 | 0.862 |
| Bagging-Random Tree | 0.896 | 0.688 | 0.742 | 0.927 | 0.896 | 0.908 | 0.827 |
| Random Committee-Random Forest | 0.912 | 0.665 | 0.728 | 0.929 | 0.912 | 0.919 | 0.863 |
| Random Committee-Random Tree | 0.885 | 0.658 | 0.721 | 0.923 | 0.885 | 0.900 | 0.832 |

The first noteworthy observation in Table 1 is the high accuracy achieved by all classifiers, which exceeds 90%. As discussed above, this fact does not ensure that the classification results for the minority class are good. This can be confirmed by examining the specific metrics for imbalanced data, G-mean and OP, whose values are quite low. The objective of sampling strategies is to increase the value of these metrics without significantly decreasing accuracy. In Tables 2 and 3, we can see that SMOTE, with both 100% and 500% resampling, hardly improves the results of OP and G-mean; in fact, these results worsen in some cases. However, the values of AUC are higher for SMOTE 100%.

In order to analyze the behavior of classifiers built from the dataset before and after applying the sampling strategies, the results of the main metrics have been represented in Figures 2–6. Our clustering-RUS approach is represented by the light blue bar. As expected, the accuracy values are lower when applying undersampling methods. However, the values of the metrics OP and G-mean, which are specific for imbalanced data, are significantly higher. This means that a better classification of the minority class instances is achieved when using undersampling strategies. In this study, the minority class, representing infected patients, is the most important because it is the target of the prediction. Therefore, an increase in minority-class precision may justify a decrease in accuracy. That is what the metrics OP and G-mean reflect; based on their values, we can deduce that SMOTE is not a good resampling approach for these data, since the results of the metrics for this strategy are very similar to the ones obtained for the original dataset.

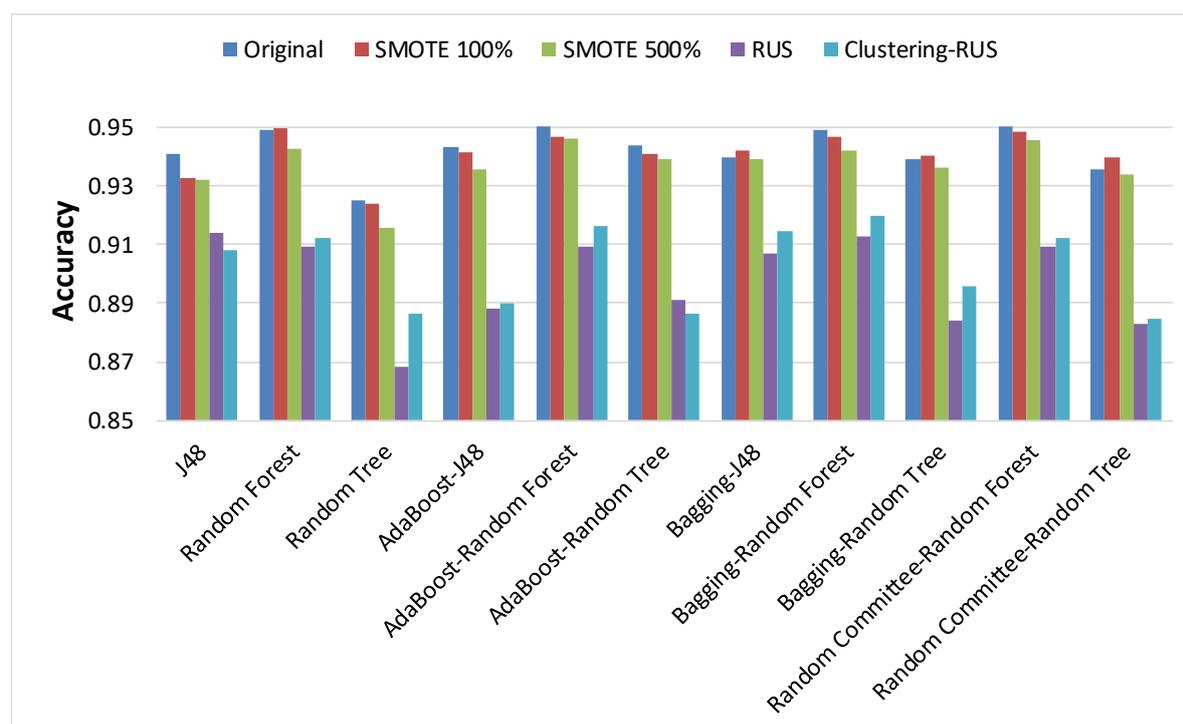

**Figure 2.** Accuracy achieved by all tested classifiers from the original and resampled datasets.

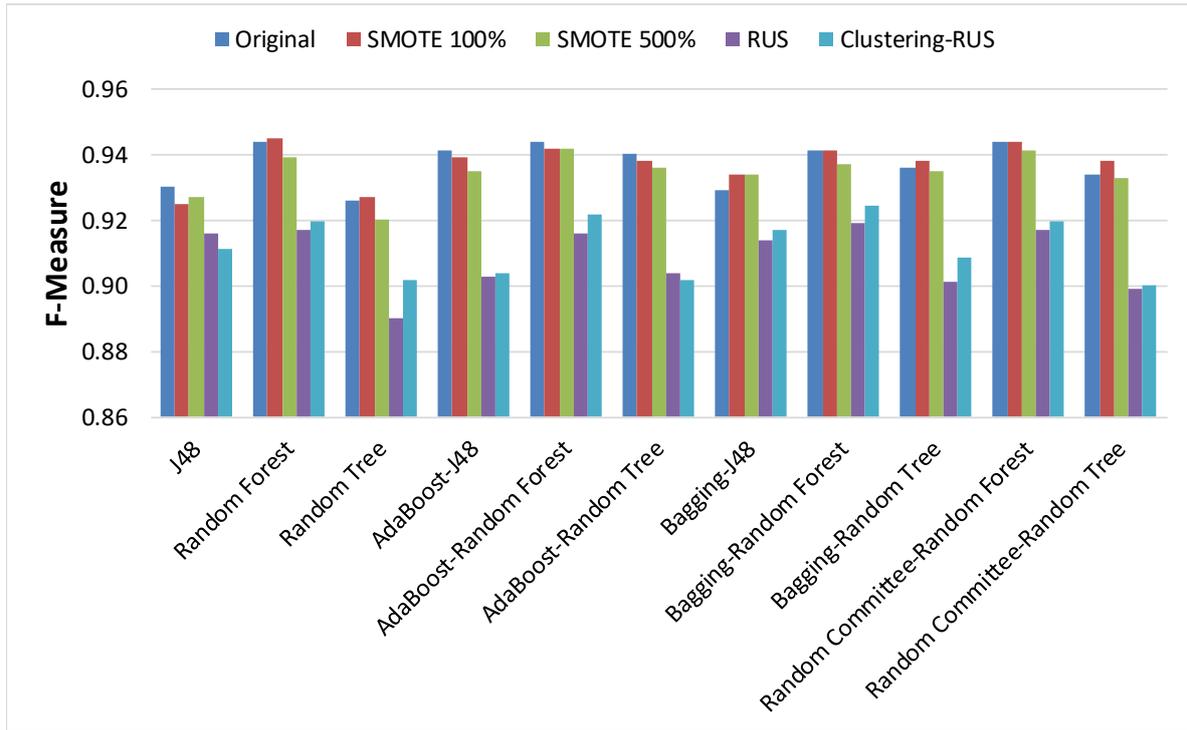

**Figure 3.** F-measure obtained by all tested classifiers from the original and resampled datasets.

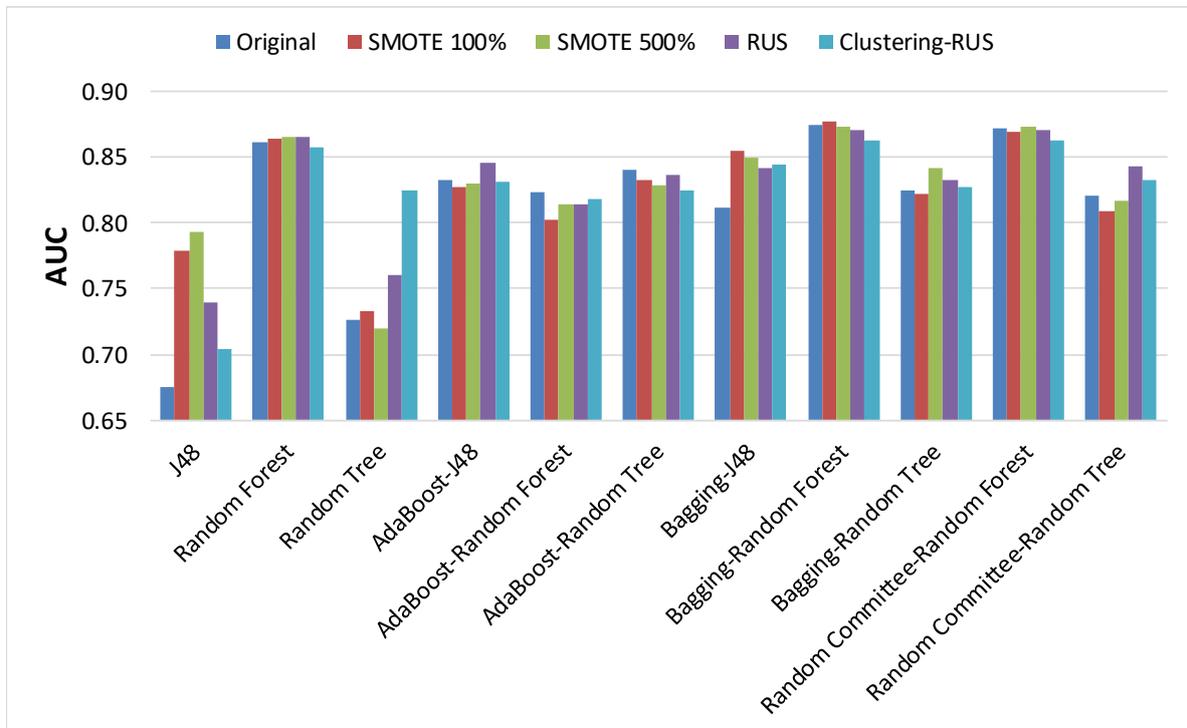

**Figure 4.** Area under the receiver operator characteristic (ROC) curve (AUC) obtained by all tested classifiers from the original and resampled datasets.

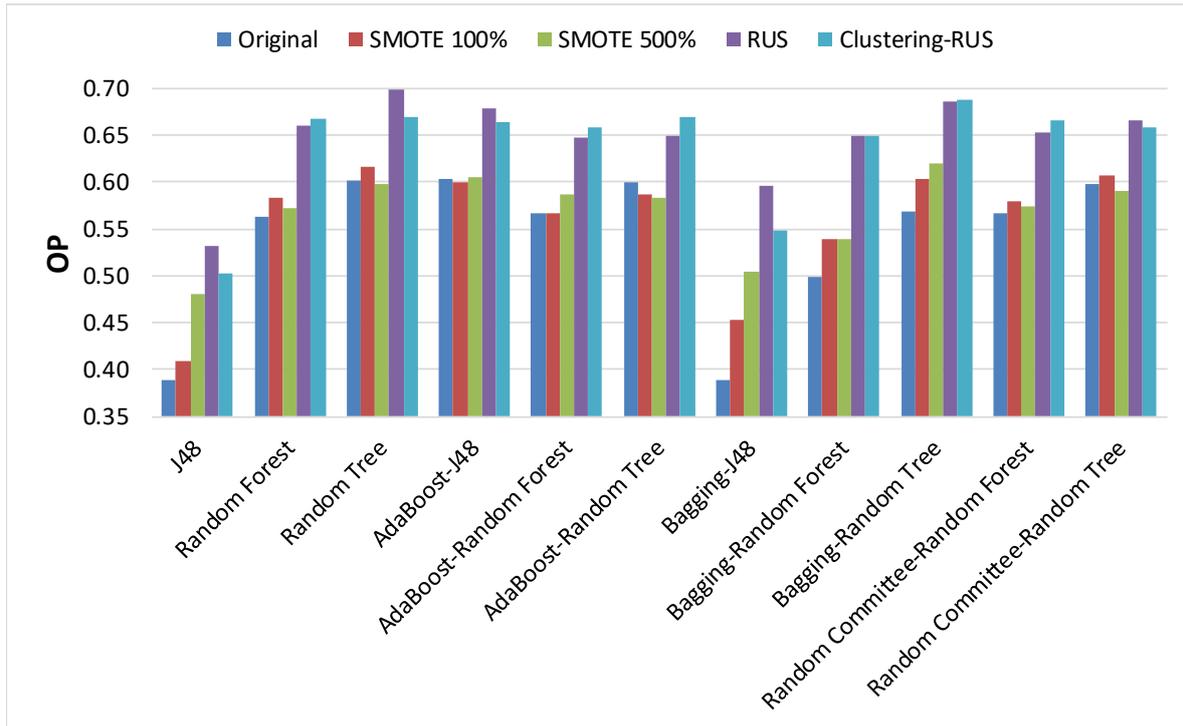

**Figure 5.** Optimized precision (OP) metric for all tested classifiers from the original and resampled datasets.

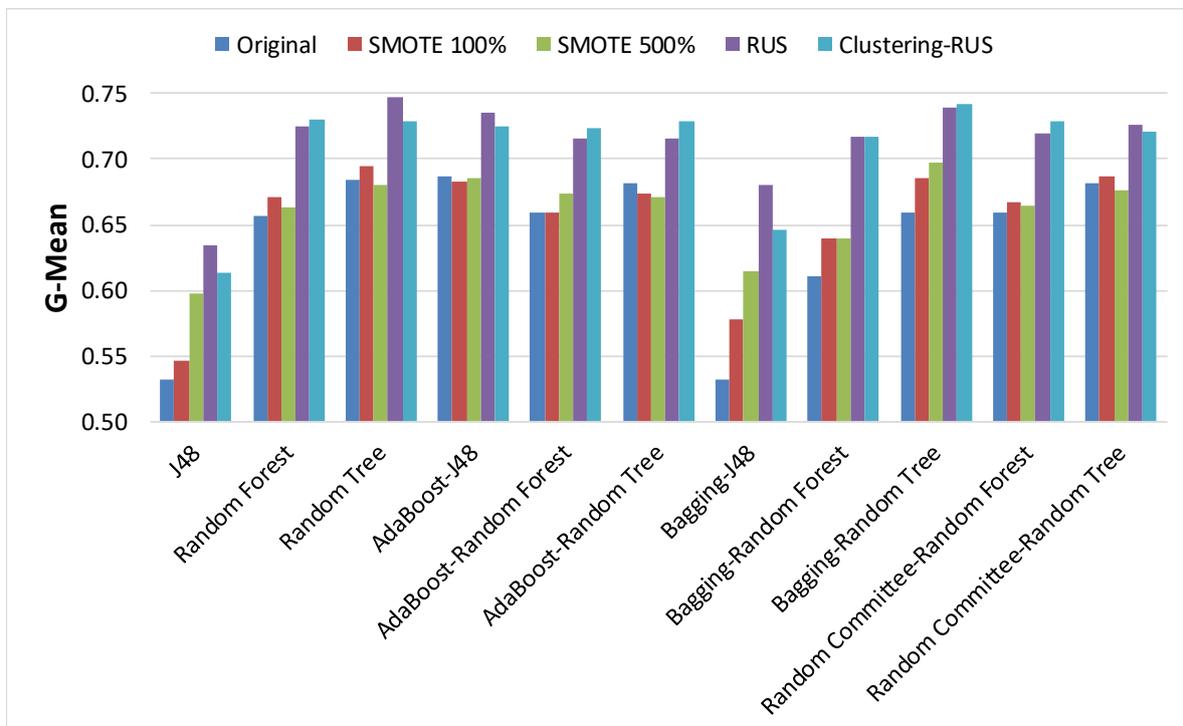

**Figure 6.** G-mean metric for all tested classifiers from the original and resampled datasets.

When comparing RUS and clustering-RUS, the values of OP and G-Mean are similar; RUS is better for some classifiers, but clustering-RUS provides better results for others. However, analyzing accuracy and F-measure, we can observe that the values for clustering-RUS are higher than the values for RUS in most of the cases. Figures 5 and 6 show that the best results of OP and G-Mean with RUS are achieved by the random tree classifier. Nevertheless, as can be seen in Figures 2 and 3, this classifier produces the lowest values in terms of accuracy and F-measure for all datasets, especially when it is applied to data undersampled with RUS. The loss of accuracy for the random tree classifier when using RUS against the original dataset is 6.16%, while the loss yielded when using clustering-RUS only amounts to 4.22%.

Regarding AUC, Figure 4 shows that there are hardly any significant differences between the resampling strategies, although there is a more uniform behavior of clustering-RUS compared to the others.

In order to compare the algorithm and to know the significance of the results for OP and G-mean (the metrics specific for imbalanced classification) when using the dataset resampled with the clustering-RUS approach, we performed the Friedman test and a post-hoc analysis based on the Wilcoxon–Holm method. The *p*-values obtained in the comparative of the algorithms for OP and G-mean, setting a significance level $\alpha = 0.05$, are given in Tables 6 and 7 respectively.

**Table 6.** *p*-values yielded by the statistical tests for the OP metric and the clustering-RUS resampling strategy.

| Algorithm | - | 2 | 3 | 4 | 5 | 6 | 7 | 8 | 9 | 10 | 11 |
|---|---|---|---|---|---|---|---|---|---|---|---|
| J48 | 1 | 0.011 | 0.011 | 0.017 | 0.011 | 0.011 | 0.017 | 0.011 | 0.011 | 0.011 | 0.011 |
| Random Forest | 2 | | 0.673 | 0.122 | 0.888 | 0.398 | 0.049 | 0.574 | 0.017 | 0.205 | 0.159 |
| Random Tree | 3 | | | 0.159 | 0.325 | 0.575 | 0.049 | 1.000 | 0.011 | 0.778 | 0.673 |
| AdaBoost-J48 | 4 | | | | 0.482 | 0.159 | 0.482 | 0.482 | 0.011 | 0.011 | 0.673 |
| AdaBoost-Random Forest | 5 | | | | | 0.159 | 0.017 | 0.205 | 0.011 | 0.122 | 0.673 |
| AdaBoost-Random Tree | 6 | | | | | | 0.024 | 1.000 | 0.011 | 0.673 | 0.673 |
| Bagging-J48 | 7 | | | | | | | 0.011 | 0.011 | 0.017 | 0.205 |
| Bagging-Random Forest | 8 | | | | | | | | 0.011 | 0.398 | 0.673 |
| Bagging-Random Tree | 9 | | | | | | | | | 0.067 | 0.011 |
| Random Committee-Random Forest | 10 | | | | | | | | | | 0.122 |
| Random Committee-Random Tree | 11 | | | | | | | | | | |

**Table 7.** *p*-values yielded by the statistical tests for the G-mean metric and the clustering-RUS resampling strategy.

| Algorithm | - | 2 | 3 | 4 | 5 | 6 | 7 | 8 | 9 | 10 | 11 |
|---|---|---|---|---|---|---|---|---|---|---|---|
| J48 | 1 | 0.018 | 0.018 | 0.042 | 0.018 | 0.018 | 0.028 | 0.018 | 0.018 | 0.018 | 0.018 |
| Random Forest | 2 | | 0.128 | 0.018 | 0.236 | 0.498 | 0.028 | 0.062 | 0.018 | 0.128 | 0.866 |
| Random Tree | 3 | | | 0.028 | 0.128 | 0.063 | 0.397 | 0.063 | 0.018 | 0.063 | 1.000 |
| AdaBoost-J48 | 4 | | | | 0.063 | 0.018 | 0.866 | 0.063 | 0.018 | 0.018 | 0.734 |
| AdaBoost-Random Forest | 5 | | | | | 0.611 | 0.028 | 0.236 | 0.018 | 0.498 | 0.866 |
| AdaBoost-Random Tree | 6 | | | | | | 0.063 | 0.397 | 0.018 | 0.128 | 0.866 |
| Bagging-J48 | 7 | | | | | | | 0.018 | 0.018 | 0.028 | 0.176 |
| Bagging-Random Forest | 8 | | | | | | | | 0.018 | 0.866 | 0.735 |
| Bagging-Random Tree | 9 | | | | | | | | | 0.018 | 0.236 |
| Random Committee-Random Forest | 10 | | | | | | | | | | 0.236 |
| Random Committee-Random Tree | 11 | | | | | | | | | | |

In addition, critical difference diagrams were used to represent the results (Figures 7 and 8). These diagrams show the classifiers ranked by performance. Considering OP and G-mean metrics, the best classifiers are the obtained with bagging-random tree and random committee-random forest, although the thick horizontal line linking all classifiers indicates that the differences between them are not significant. However, it can be observed in the tables that the lowest p-values are obtained when J48 and bagging-random tree are compared with the rest of the algorithms, that is, the greatest differences are obtained for the worst and the best algorithm with respect to the rest.

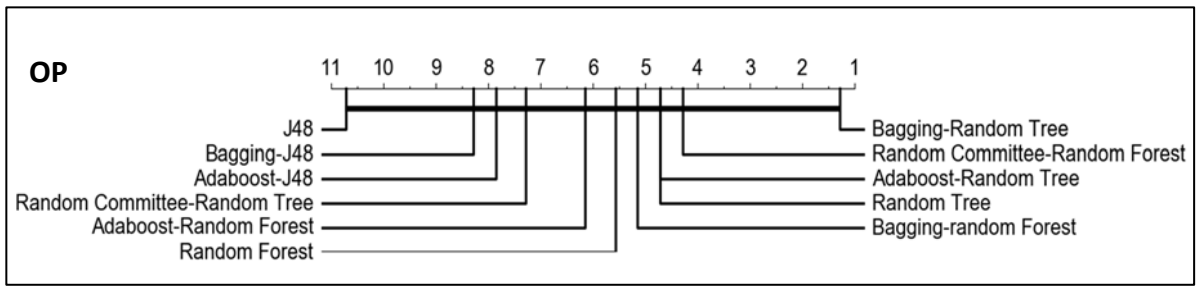

**Figure 7.** Critical difference diagram to compare the classifiers regarding the OP metric for the clustering-RUS resampling strategy.

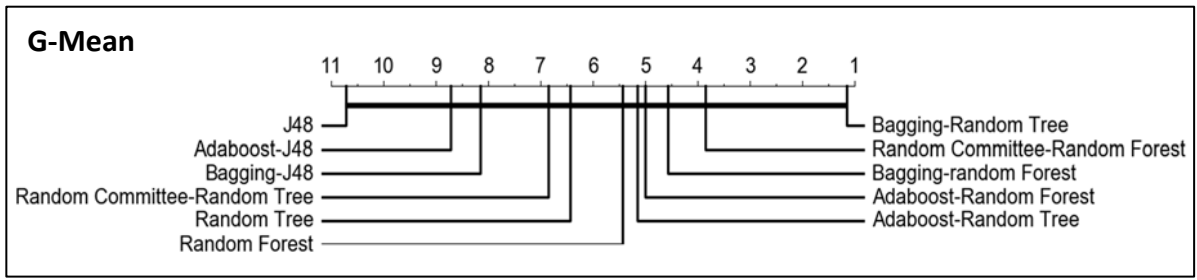

**Figure 8.** Critical difference diagram to compare the classifiers regarding the G-mean metric for the clustering-RUS resampling strategy.

Since the objective of this work is to prove the improvement of the algorithm performance when using our clustering-based resampling proposal, we have applied the same significance tests to compare the four best classifiers induced from the datasets resampled with both RUS and clustering-RUS, which are the resampling techniques that have shown the best behavior in the experiments. Figures 9 and 10 show the results, which evidence the better performance of all the algorithms with clustering-RUS (C-RUS in the figures) vs. RUS.

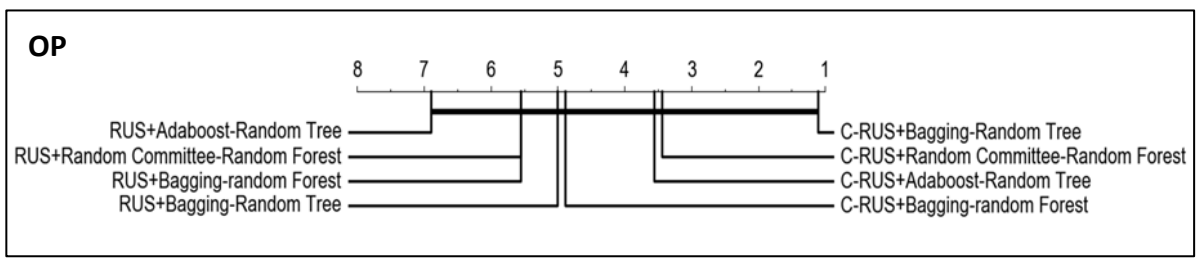

**Figure 9.** Critical difference diagram to compare the classifiers obtained using RUS vs. clustering-RUS (C-RUS) regarding the OP metric.

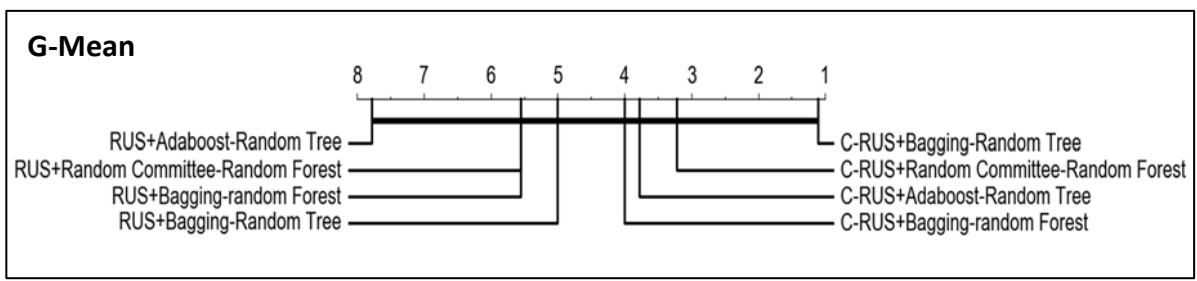

**Figure 10.** Critical difference diagram to compare the classifiers obtained using RUS vs. clustering-RUS (C-RUS) regarding the G-mean metric.

Another conclusion that can be deduced from the graphs in Figures 2–6 is that the increase in the values of OP and G-mean for classifiers used with RUS and clustering-RUS is accompanied by the decrease in the values of accuracy and F-measure. Therefore, it would be interesting to find the classifier-sampling method pair that provides the best balance between those metrics.

For this purpose, the highest accuracy value and the highest OP value achieved in the experiments have been identified, and the percentage of difference (loss) with respect to these maximum values has been calculated for all the algorithm-sampling strategy pairs. Finally, the differences for the two metrics have been averaged. The pairs with the lowest average loss values will be the most appropriate, as they are the ones that provide the lowest global loss between accuracy and OP. The same procedure has been carried out for accuracy and G-mean.

Table 8 shows the accuracy values for all the algorithm-sampling method pairs, as well as the loss of accuracy of all those pairs with respect to the best value. The best accuracy is 0.92, which is provided by the algorithm bagging-random forest and the clustering-RUS sampling method.

The values of OP and G-mean and their corresponding loss with respect to the best value (OP = 0.70 and G-Mean = 0.75) are presented in Tables 9 and 10. The last columns of these tables contain the average loss of accuracy and OP/G-Mean, respectively. This information is also represented in Figures 11 and 12, with the aim of better visualizing the lowest values and the differences between the use of RUS and clustering-RUS.

**Table 8.** Accuracy and loss of accuracy with respect to the best value of accuracy (0.92) for the algorithm-sampling method pairs.

| Algorithm | Accuracy | | % Accuracy Loss with Respect to the Best Value | |
|---|---|---|---|---|
| | RUS | Clustering-RUS | RUS | Clustering-RUS |
| J48 | 0.91 | 0.91 | 0.65 | 1.26 |
| Random Forest | 0.91 | 0.91 | 1.20 | 0.84 |
| Random Tree | 0.87 | 0.89 | 5.65 | 3.64 |
| AdaBoost-J48 | 0.89 | 0.89 | 3.48 | 3.24 |
| AdaBoost-Random Forest | 0.91 | 0.92 | 1.20 | 0.42 |
| AdaBoost-Random Tree | 0.89 | 0.89 | 3.15 | 3.64 |
| Bagging-J48 | 0.91 | 0.91 | 1.41 | 0.61 |
| Bagging-Random Forest | 0.91 | 0.92 | 0.76 | 0.00 |
| Bagging-Random Tree | 0.88 | 0.90 | 3.91 | 2.65 |
| Random Committee-Random Forest | 0.91 | 0.91 | 1.20 | 0.84 |
| Random Committee-Random Tree | 0.88 | 0.88 | 4.00 | 3.83 |

**Table 9.** Loss of OP with respect to the best value of OP (0.70), and average loss of accuracy and OP for the algorithm-sampling method pairs.

| Algorithm | OP | | % OP Loss with Respect to the Best Value | | % Average Loss of Accuracy and OP | |
|---|---|---|---|---|---|---|
| | RUS | Clustering-RUS | RUS | Clustering-RUS | RUS | Clustering-RUS |
| J48 | 0.53 | 0.50 | 24.05 | 28.32 | 12.35 | 14.79 |
| Random Forest | 0.66 | 0.67 | 5.55 | 4.58 | 3.37 | 2.71 |
| Random Tree | 0.70 | 0.67 | 0.05 | 4.26 | 2.85 | 3.95 |
| AdaBoost-J48 | 0.68 | 0.66 | 2.97 | 5.21 | 3.22 | 4.23 |
| AdaBoost-Random Forest | 0.65 | 0.66 | 7.54 | 5.91 | 4.37 | 3.16 |
| AdaBoost-Random Tree | 0.65 | 0.67 | 7.22 | 4.26 | 5.18 | 3.95 |
| Bagging-J48 | 0.60 | 0.55 | 14.78 | 21.62 | 8.10 | 11.11 |
| Bagging-Random Forest | 0.65 | 0.65 | 7.26 | 7.27 | 4.01 | 3.63 |
| Bagging-Random Tree | 0.69 | 0.69 | 2.10 | 1.75 | 3.00 | 2.20 |
| Random Committee-Random Forest | 0.65 | 0.67 | 6.82 | 4.97 | 4.01 | 2.91 |
| Random Committee-Random Tree | 0.67 | 0.66 | 4.86 | 5.95 | 4.43 | 4.89 |

**Table 10.** Loss of G-mean with respect to the best value of G-Mean (0.75), and average loss of accuracy and G-mean for the algorithm-sampling method pairs.

| Algorithm | G-Mean | | % G-Mean Loss with Respect to the Best Value | | % Average Loss of Accuracy and G-Mean | |
| --- | --- | --- | --- | --- | --- | --- |
| | RUS | Clustering-RUS | RUS | Clustering-RUS | RUS | Clustering-RUS |
| J48 | 0.63 | 0.61 | 15.05 | 17.79 | 7.85 | 9.53 |
| Random Forest | 0.73 | 0.73 | 2.89 | 2.21 | 2.04 | 1.53 |
| Random Tree | 0.75 | 0.73 | 0.00 | 2.36 | 2.83 | 3.00 |
| AdaBoost-J48 | 0.74 | 0.72 | 1.51 | 2.92 | 2.49 | 3.08 |
| AdaBoost-Random Forest | 0.72 | 0.72 | 4.22 | 3.05 | 2.71 | 1.73 |
| AdaBoost-Random Tree | 0.72 | 0.73 | 4.22 | 2.36 | 3.69 | 3.00 |
| Bagging-J48 | 0.68 | 0.65 | 8.98 | 13.42 | 5.20 | 7.02 |
| Bagging-Random Forest | 0.72 | 0.72 | 4.02 | 3.91 | 2.39 | 1.96 |
| Bagging-Random Tree | 0.74 | 0.74 | 0.90 | 0.56 | 2.41 | 1.61 |
| Random Committee-Random Forest | 0.72 | 0.73 | 3.70 | 2.46 | 2.45 | 1.65 |
| Random Committee-Random Tree | 0.73 | 0.72 | 2.81 | 3.49 | 3.40 | 3.66 |

In Table 9, we can see that the highest OP value, namely, 0.70, is obtained with the random tree algorithm and the RUS sampling method. However, this pair cannot be considered the best, since it also provides the lowest accuracy value, as shown in Figure 2 and Table 8. The best pair will be the one that provides a good balance between accuracy and OP. In our case, bagging-random tree, used together with clustering-RUS, would be the best, providing the lowest average loss of accuracy and OP with respect to the best values of those metrics. Table 9 shows that this average loss only amounts to 2.20%, which is far from the worst average loss of 14.79%.

The same behavior can be observed when analyzing G-mean values versus accuracy (Table 10). Once more, the best value of G-mean, 0.75, is yielded by random tree paired with RUS, but the lowest average loss of accuracy and G-mean is given by the pair bagging-random tree and clustering-RUS, with a value of 1.61%, followed very closely by random committee-random forest and clustering-RUS, with 1.65%. In all cases, our clustering-RUS proposal is involved in the best pairs.

Figures 11 and 12 also show that our sampling strategy provides better results than the simple RUS strategy for almost all algorithms. They also show the bad behavior of the J48 algorithm, both when used as a simple classifier and as part of ensemble classifiers.

In order to corroborate the conclusions drawn from the tables above, Figure 13 represents the success rates of the minority class (TPR) and those of the majority class (TNR). It is well known that the increase in the former always occurs at the expense of the latter. With this in mind, the first thing that emerges from the examination of the graph is the significant increase in TPR when using undersampling in comparison with the small improvement achieved when using the oversampling strategy. In addition, the increase obtained from undersampling occurs at the cost of a very small decrease in the correct classification rate of the majority class (TNR).

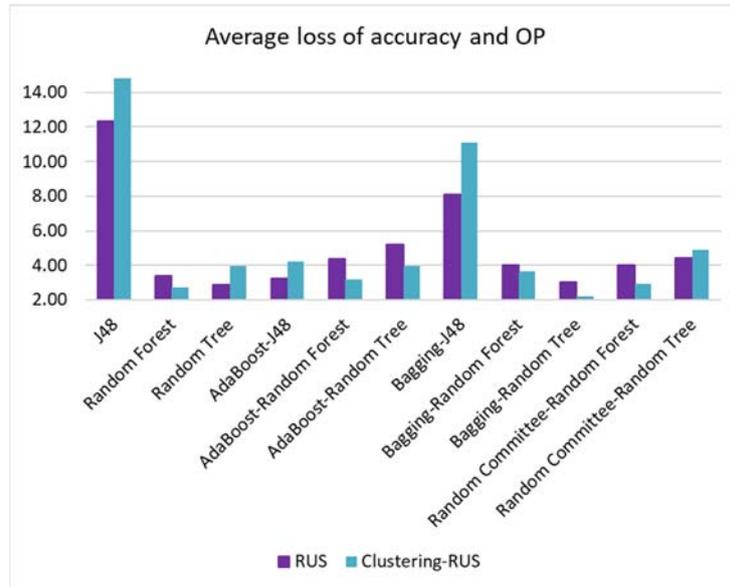

**Figure 11.** Average loss of accuracy and OP with respect to the best values of each metric.

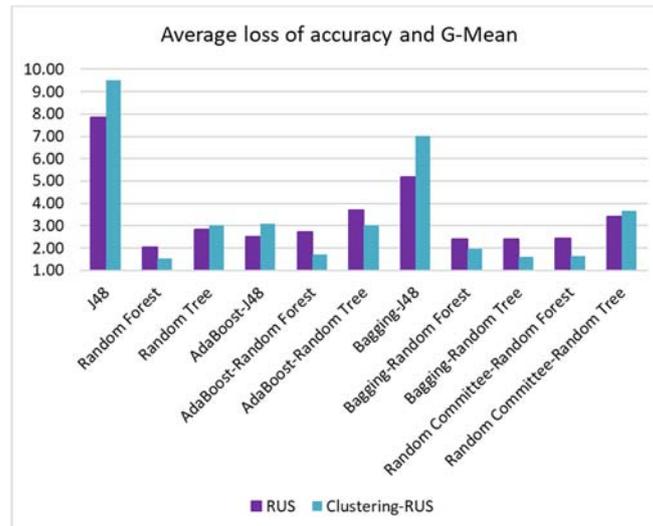

**Figure 12.** Average loss of accuracy and G-mean with respect to the best values of each metric.

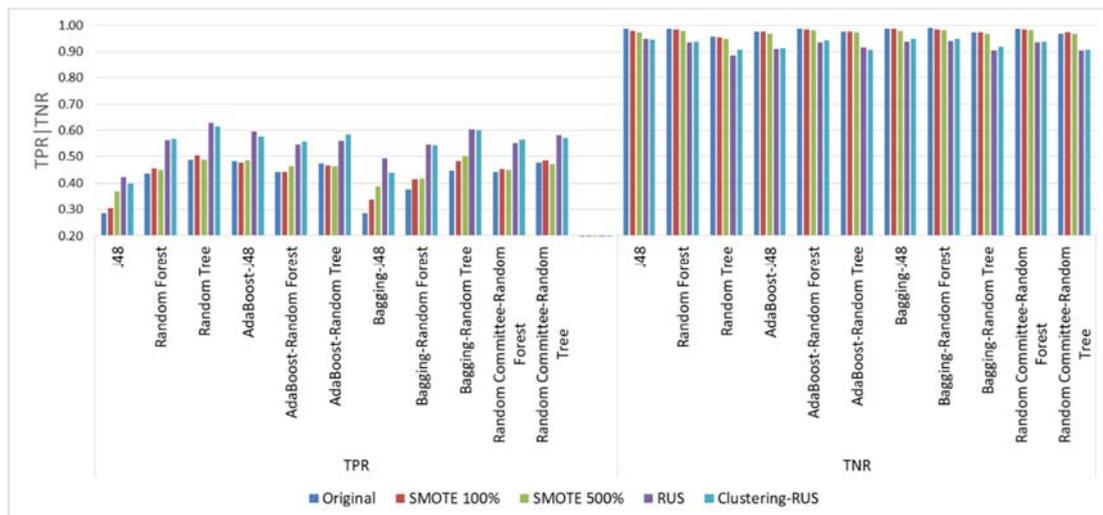

**Figure 13.** True positive rate (TPR) and true negative rate (TNR) for all tested algorithms and sampling strategies.

The graph in Figure 13 also corroborates that bagging-random tree used with clustering-RUS provides the best TPR values and the lowest decrease in TNR.

*4.3. Detection of the Most Influential Factors*

Although the purpose of this work is to propose a method for obtaining reliable predictive models from imbalanced datasets in the specific HAI application domain, we have expanded the study in order to obtain deeper knowledge from the medical point of view. To this end, some feature selection methods have been applied to determine which are the most influential factors in infection acquisition. We have considered both methods based on the gain of information provided by the feature with respect to the class [54] and the CFS (correlation-based feature subset selection) approach [55] which also considers the correlation between the factors or features. CFS evaluates the significance (merit) of a subset of features (attributes), taking into account the individual predictive ability of each feature and the degree of redundancy between them. This method selects the subsets of attributes that are highly correlated with the class while having low inter-correlation between them.

The method of information gain ratio yielded the following ranked list of attributes, where the values in parentheses are the ratios of information gain: extrarenal depuration (0.115), parenteral nutrition (0.093), emergency surgery (0.030), neutropenia (0.028), APACHE (0.023), immunodeficiency (0.018), central venous catheter (0.016) and mechanical ventilation (0.016). Through the CFS technique, 89 subsets of attributes were automatically evaluated. The best of which, with a merit of 0.14, was formed by the following attributes: neutropenia, central venous catheter, parenteral nutrition and extrarenal depuration. Although gain ratio and merit values range from 0 to 1, values obtained in the context of this study are common in classification problems. Analyzing the output of the methods, we can observe a coincidence in 4 factors, although there is no exact coincidence in the order of importance of these features. These are: neutropenia, central venous catheter, parenteral nutrition and extrarenal depuration. This indicates that the presence of invasive devices is one of the major causes of infection in an ICU.

In addition, we have examined some models generated by the machine learning algorithms. Figure 14 shows some rules extracted from a decision tree classifier. This type of model is formed by nodes where attributes are tested, branches for the different values of the attributes and leaves for the classes, one in each leaf. The tree represents a set of rules that are checked when an instance needs to be classified. In Figure 14, the nested conditions are shown, and dashed lines are used to delimit the levels. For example, the first rule of the tree is the following: IF extrarenal depuration = N and Parenteral nutrition = Y and APACHE ≤ 17 THEN Infection = NO. This portion of the model contains 11 rules. We can also see that the attributes in the top levels of the tree (the best) are some of the selected by the feature selection methods discussed above.

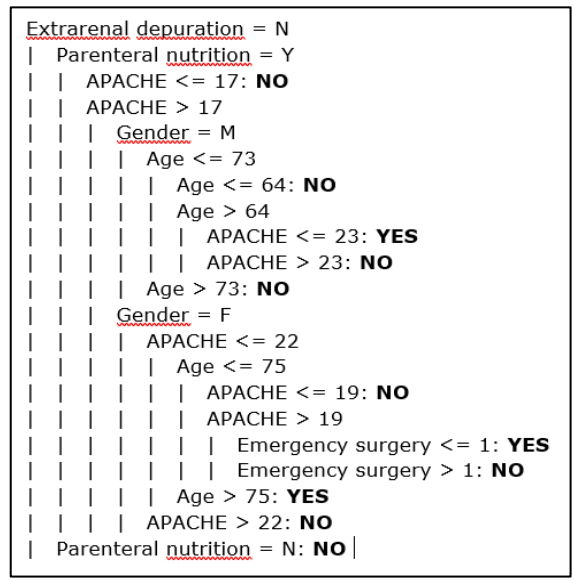

**Figure 14.** Portion of a decision tree model.

## 5. Discussion

The previous sections presents a detailed comparative analysis of the algorithms and resampling strategies tested in this work. In this section, we will turn to the most important findings regarding the reliability of the classification.

The results obtained in the experimental study conducted in the context of predictive modeling of healthcare-associated infections confirm the fact that the behavior of classification algorithms and the effectivity of resampling strategies are highly dependent on the application domain and the characteristics of the datasets. In this sense, the most remarkable aspect found in this study is that algorithms such as SVM or Bayesian networks and resampling strategies such as SMOTE yield poor results in the context of our work despite being effective in other fields.

As expected, the results also show that resampling methods, which are used to address the problem of imbalanced data classification, improve precision in the classification of the minority class examples at the cost of global accuracy. Therefore, the target is to find the method that provides the best balance between those aspects. Most of the work in the literature only evaluates classic metrics such as accuracy, precision, recall and AUC, but, as shown in the empirical study, those metrics are not suitable to identify the classifiers that ensure the best balance. We have computed the values of two additional metrics, G-mean and OP, specifically designed for imbalanced data classification. Their values reveal a significant better behavior of RUS and clustering-RUS sampling strategies against SMOTE for all of the tested classifiers. The statistical significance tests also showed the superiority of clustering-RUS vs. RUS in terms of G-mean and OP for the best classifiers. Although the difference was not significant, the values of accuracy and F-measure were better for clustering-RUS. This indicates that the latter achieves the best balance between the rate of improvement in the classification of the minority class examples and the loss of accuracy. This fact is further corroborated by the subsequent analysis of all classifiers, in which we studied the loss in accuracy, OP and G-mean values with respect to the best value of each, and we evaluated the combined loss of both OP and accuracy, and G-mean and accuracy. The lowest values of combined lost were yielded by the clustering-RUS sampling strategy and the bagging-random tree classifier. This, in turn, shows that an ensemble classifier (bagging) accompanied by a base classifier that is also an ensemble (random tree) are the most suitable for making predictions in the scope of our work.

In addition, TPR and TNR were analyzed taking into account that the minority class is the positive. It is known that the increase in TPR produces a decrease in TNR. However, when using RUS and clustering-RUS, we found that the TPR values provided by the algorithms tested are significantly higher than those obtained when using SMOTE. By contrast, the TNR values for RUS and clustering-RUS are slightly lower than for SMOTE. The highest values of TPR were obtained for the algorithms

*Appl. Sci.* **2019**, *9*, 5287　　　　　　　　　　　　　　　　　　　　　　　　　　　　　　　　　　　24 of 27random tree and bagging-random tree with RUS and clustering-RUS; especially when clustering-RUS was used, bagging-random tree achieved a higher value of TNR than random tree.

Given the fact that the objective is to improve the classification reliability of the minority class instances without significantly decreasing the classification reliability of the majority class instances and accuracy, we can say that clustering-RUS is the most suitable approach for dealing with imbalanced data in the context of our study. The study also proves our initial hypothesis that ensemble classifiers present a better behavior than single classifiers in these situations. However, although better results have been obtained in this context with the proposed resampling method and with the use of a certain ensemble classifier, it is necessary to treat the findings with caution since they may not be extensible to other application domains. As proved in several previous studies, the effectiveness of resampling techniques is not only sensitive to the imbalance ratio of datasets but to many other data characteristics. The same applies to ensembles, whose behavior varies considerably from one domain to another.

## 6. Conclusions

In this work, the data from 4616 patients hospitalized in an ICU have been processed with different data-mining algorithms in an attempt to induce models for predicting healthcare-associated infections in future patients. It is necessary to address the problem of building classifiers from imbalanced datasets because only 6.7% of the patients included in the dataset presented infections, while 93.3% did not contract any infection whatsoever.

Our proposal is focused on the application of ensemble classifiers and an undersampling strategy that takes into account the imbalance ratio in different regions in the space of characteristics. The purpose is to avoid problems such as overfitting and loss of information, derived from the use of resampling strategies, as well as certain difficulties in the implementation of other approaches, such as algorithm modification and cost-sensitive learning.

In order to validate the proposal, several algorithms have been tested using both original and resampled datasets. Apart from our proposal, the SMOTE oversampling approach and the RUS strategy have also been evaluated.

The results of the metrics used to assess the performance of the algorithms and the sampling strategies showed that the best balance between accuracy and the values of OP and G-mean is yielded by the bagging ensemble with random tree as base classifier when it is applied to the dataset undersampled by our clustering-RUS proposal. This algorithm with the clustering-RUS resampling method yielded a 2.2% average loss of accuracy and OP versus 14.79% yielded by the J48 decision tree, the worst classifier. Bagging-random tree with RUS gave a higher loss, 3.3%. The average loss of accuracy and G-mean was 1.61% for bagging-random tree versus 9.53% for J48. For RUS, when using bagging-random tree, this loss was also higher, 2.45%. These data and the statistical tests conducted in this study also highlighted the generally better behavior of the proposed clustering-RUS sampling method compared to the RUS approach. In addition, the experiments revealed that the SMOTE oversampling method delivers poor results in the context of this work, in spite of it being one of the most widely used approaches to address imbalanced data classification.

**Author Contributions:** Conceptualization, resources, methodology and validation, F.S.-H. and J.C.B.-H.; formal analysis, investigation, M.S.-B.; data curation, software, investigation, M.S.K.; methodology, supervision, writing—review and editing, M.N.M.-G.

**Funding:** This research received no external funding.

**Conflicts of Interest:** The authors declare no conflict of interest.## References

1. Haque, M.; Sartelli, M.; McKimm, J.; Bakar, M.A. Health care-associated infections—An overview. *Infect. Drug. Resist.* **2018**, *11*, 2321–2333. doi:10.2147/IDR.S177247.
2. Scott, R.D.; Culler, S.D.; Rask, K.J. Understanding the Economic Impact of Health Care-Associated Infections: A Cost Perspective Analysis. *J. Infus. Nurs.* **2019**, *42*, 61–69. doi:10.1097/NAN.0000000000000313.